# ELIMINATION OF GLASS ARTIFACTS AND OBJECT SEGMENTATION

Vini Katyal
Amity School of Engineering
Amity University
Uttar Pradesh
India
vini_katyal@yahoo.co.in

Aviral
Amity School of Engineering
Amity University
Uttar Pradesh
India
kaviral12@yahoo.com

Deepesh Srivastava
Amity School of Engineering
Amity University
Uttar Pradesh
India
dksrivastava@amity.edu

## ABSTRACT

Many images nowadays are captured from behind the glasses and may have certain stains discrepancy because of glass and must be processed to make differentiation between the glass and objects behind it. This research paper proposes an algorithm to remove the damaged or corrupted part of the image and make it consistent with other part of the image and to segment objects behind the glass. The damaged part is removed using total variation inpainting method and segmentation is done using kmeans clustering, anisotropic diffusion and watershed transformation. The final output is obtained by interpolation. This algorithm can be useful to applications in which some part of the images are corrupted due to data transmission or needs to segment objects from an image for further processing.

## 1. INTRODUCTION

Object segmentation is widely used in object tracking and recognition. The approaches of object segmentation can be classified into 3 categories [1]: global knowledge based, region based and edge based. Most methods deal with only segmentation and don't pre process the image to remove unwanted objects that might hinder the efficiency of the segmentation. Many techniques have been developed for small object segmentation these include Visual attention based small object segmentation in natural images which narrows the searching region to increase the segmentation accuracy [2], Adaptive approach to small-object segmentation which has used genetic algorithm for segmentation [3], Many new methods utilize a combination of K-means clustering and watershed for better object segmentation and to avoid over segmentation by watershed Image Segmentation with Clustering K-Means and Watershed Transform. In 2000 an algorithm was proposed that segments an object from the background of a scene where the scene is illuminated by unknown ambient light source but it didn't solve the problem of disturbances and unwanted material on the glass when the image was taken [4]. Another paper targeted the separation of reflections and lighting using independent components analysis [5].Thus all the existing methods consider only one problem at a time either that of image in-painting or small object segmentation. This paper combines the ideologies of both. The paper is organized as follows: Section 2 describes the steps of the proposed algorithm Section 3 presents the experimental results with more details about the procedure adopted Finally Section 4 presents our conclusions.

## 2. PROPOSED ALGORITHM

This paper proposes the segmentation of objects behind the glass window and also removal of any unwanted stain and scratches on the window which can hinder the process of detection of objects. This technique is suited for images which are captured from a distance behind the glass. In this algorithm we first remove the unwanted disturbances on the glass using total variation in-painting. The occluded area is blurred out of the original image. The next step processes the gradient magnitude which computes the amplitude of sudden change in grey level values of pixels. Following this step, anisotropic diffusion is applied to enhance the contrast and smoothes out the unwanted object in the image. The algorithm uses K-means clustering and watershed transformation techniques to segment out the objects of interest. In the end, Smooth interpolation is performed using the original gray image and the image produced after watershed transformation.

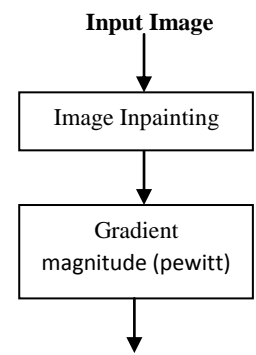

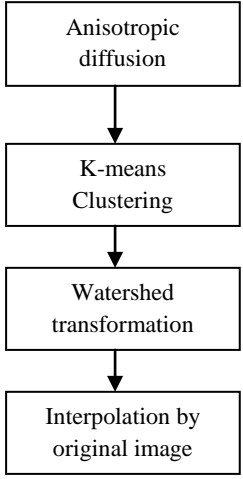

Figure 1: The proposed algorithm

## 2.1 Gradient magnitude operator

A gradient magnitude operator detects the amplitude edges at which pixel change their gray levels suddenly. For an image volume f(x), the magnitude of the gradient vector:

$$|\nabla f| = \sqrt{\left(\frac{\partial f}{\partial x}\right)^2} + \sqrt{\left(\frac{\partial f}{\partial y}\right)^2} + \sqrt{\left(\frac{\partial f}{\partial z}\right)^2} \qquad (1)$$

assumes a local maximum at an amplitude edge. The magnitude is zero, if f is constant.

In case of digital images, partial derivatives are conveniently computed by finite difference approximations. In this paper we have used 'prewitt' operator for computing the gradient magnitude [6].

## 2.2 Total variation inpainting

The total variation inpainting model is a method which concurrently fills the corrupted or spoilt region with the available information and removes noise. It has many benefits as for easy implementation, edge preserving abilities and removing noise. It also serves as a basic building block for other inpainting algorithms.

Total variation in-painting model demands a balance between a fidelity term and a fitting term. The fitting term is only applied to the non-occluded regions and the fidelity term is applied on the whole image. Precisely

$$E_{inpaint}(u) = \alpha/2 \int_{\vartheta/D_i} (u-f)^2 dx + \int_{D_i} |\nabla u|\, dx \qquad (2)$$

Where $\vartheta$ is the entire image domain and $D_i$ is the in-painting domain [7].

## 2.3 Anisotropic diffusion for edge enhancement

Anisotropic Diffusion is used in our algorithm to enhance the image and smoothing the edge of the objects. It is a useful technique to perform segmentation of objects in an image. This paper has made use of the coefficient which offers edge preservation while diffusion takes place [8].

Let $\Omega \rightarrow R^2$ denote a subset of plane and $I(.,t):\Omega \rightarrow R$ be a family of gray scale images, and then anisotropic diffusion is defined by:

$$\frac{\partial I}{\partial t} = div(c(x,y,t)\nabla I) = \nabla c \cdot \nabla I + c(x,y,t)\Delta I \qquad (3)$$

Where $\Delta$ denotes the laplacian, div (..) is the divergence operator, and c(x,y,t) is the diffusion coefficient. It also controls the rate of diffusion according to the orientation of filter matrix during diffusion. The edge preserving coefficient can be defined as follows:

$$C(||\Delta I||) = e^{-\left(\frac{||\Delta I||}{k}\right)^2} \qquad (4)$$

Diffusion is applied in the north, south, east and west directions, anisotropic diffusion is suited for edge enhancement and stops diffusion across edges. It also serves the purpose of smoothening.

## 2.4 K-means clustering

K-means is the simplest unsupervised learning algorithm where the given data or image in our case is classified into a number of clusters which is found to be 5 in the algorithms case. Then k centroids are defined for each cluster. The next step is to take each point belonging to a given data set and associate it to the nearest centroid. When no point is left the first step is complete at this point we need to re-calculate k new centroids as bary centers of the clusters resulting from the previous step. After we have these k new centroids, a new binding has to be done between the same data set points and the nearest new centroid. A loop has been generated [9].

The algorithm aims at minimizing an objective function, in this case a squared error function. The objective function:

$$J = \sum_{j=1}^{K}\sum_{i=1}^{N} ||x_i^{(j)} - c_j||^2 \qquad (5)$$

Where:

$||x_i^{(j)} - c_j||^2$ Is chosen distance measured between the data $x_i^{(j)}$ and the cluster centroid $c_j$;

The algorithm has the following steps:

- We choose the number of clusters, K;
- We then randomly chose K pixels representing the initial group centroids;
- We assign each pixel to the group that has the closest centroid;
- When all pixels have been assigned, we recalculate the positions of the K centroids;
- Repeat Steps 2 and 3 until the centroids no longer move. This produces a separation of the pixels into groups from which the metric to be minimized can be calculated.

The K-means has been used for efficient segmentation and classification of objects in the images taken however its speed could be increased by using hierarchical clustering or dynamically deciding the number of clusters

## 2.5 Watershed Transform

The watershed transform has been used for image segmentation since a very long time and is a very powerful tool for image segmentation however when used individually may lead to over segmentation sometimes. The watershed transform can be classified as a region-based segmentation approach. The intuitive idea underlying this method comes from geography: it is that of a landscape or topographic relief which is flooded by water, watersheds being the divide lines of the domains of attraction of rain falling over the region [10].

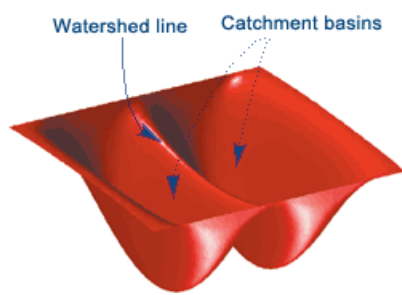

Figure 2: Watershed transform

The watershed transform is used after applying k-means clustering firstly to avoid over segmentation which took place in the image otherwise and to further efficiently segment the objects present in the images that we used. This step was necessary as our objects were small and otherwise not very distinct.

The major advantages of watershed are:

- It produces closed contours: to each minimum or to each marker corresponds one region.
- Flooding a topographic surface fills some minima, and the watershed of the flooded surface has less catchment basins. The catchment basins of successive flooding form a hierarchical segmentation.
- It is possible to flood a surface so as to impose minima at some predetermined places: this leads to marker based segmentation. [11]

Watershed when individually without k-means clustering does not produce good results in case of objects with very small size present at a distance. The efficiency of segmentation is found to increase when Watershed and k-means are used together.

## 2.6 Image Interpolation by using the original image

The mask produced is smoothly interpolated inward from the pixel values on the boundary of the polygon by solving Laplace's equation. The boundary pixels are not modified. The final output obtained through Watershed transformation is being superimposed or interpolated on the original image so as to obtain an image with distinct objects as in our case. If there are multiple regions, Image interpolation performs the interpolation on each region independently (MATLAB 7.1 help).

Laplace's equation for interpolation is given by:

$$\Delta \varphi = 0 \ Or \ \Delta^2 \varphi = 0 \qquad (5)$$

## 3. EXPERIMENTAL RESULTS

The images taken for experimental results consist of images with stains on the window and objects to be detected across the window. The steps are applied one at a time to remove stains by using Total variation in-painting, subsequently the gradient magnitude is produced using pewitt operator, the anisotropic diffusion is applied to smoothen the unwanted components with an edge preserving coefficient, finally k-means clustering and watershed our applied for segmentation and interpolation is done on the output with the original image to bring out the objects in the scene. All the images taken have some sort of stains on windows that prevent the proper segmentation of objects otherwise.

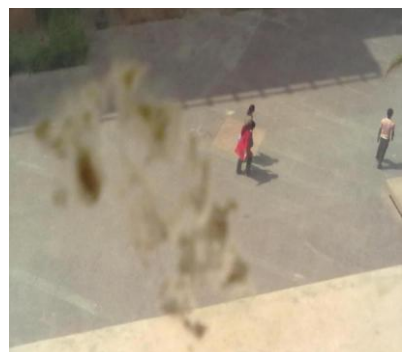

(a)

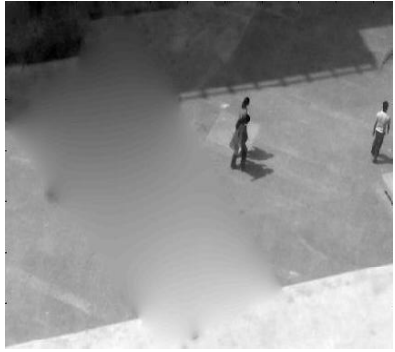

(b)

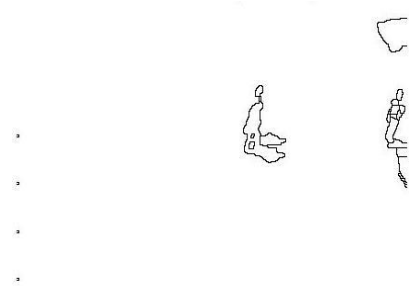

(f)

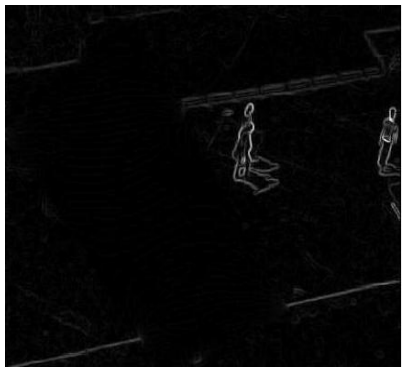

(c)

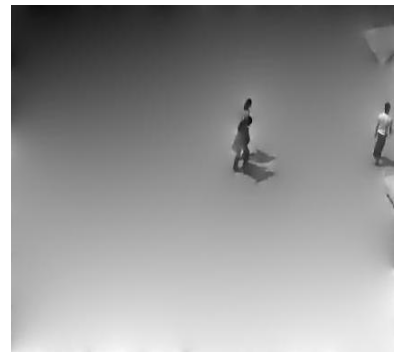

(g)

Figure 3: (a) Original Image (b) Total variation inpainting (c) Gradient magnitude (d) Anisotropic diffusion (e) K means clustering (f) Watershed transformation (g) Interpolation by original image

## 4. CONCLUSION AND FUTURE WORK

This algorithm has proposed a fast way of segmenting small objects from a natural image seen through a window plus the removal of unwanted and undesired objects from the scene which might hamper the segmentation process otherwise. This algorithm could be utilized for video sequences and CCTV's for removal of unwanted disturbances and to facilitate better object segmentation. The use of k-means clustering and watershed together lowers the chances of over segmentation. This can be extended to real time video object segmentation where chances of disturbance and unwanted materials entering are high. The images can also be converted into 3D and divided into different focal planes, the focal plane with the glass defect can be removed and the object behind the glass disturbance can be segmented efficiently

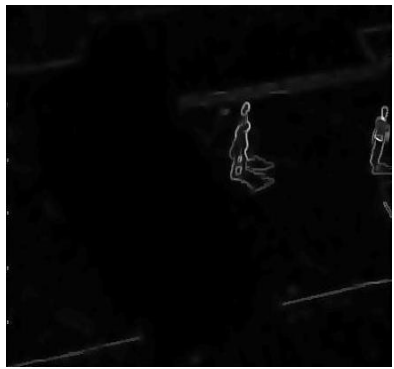

(d)

## 5. ACKNOWLEDGMENTS

The Authors wish to thank Amity University for its support in our research work.

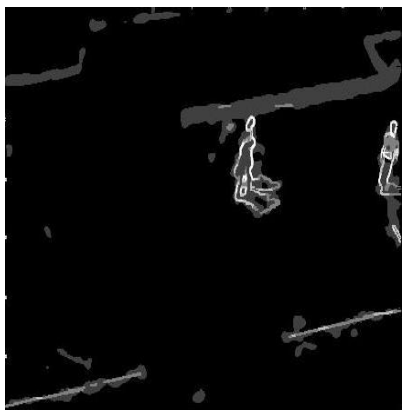

(e)